\documentclass[runningheads]{llncs}
\usepackage{cite}
\usepackage{amsmath,amssymb,amsfonts}
\usepackage{algorithmic}
\usepackage{multirow}
\usepackage{subcaption}
\usepackage{longtable}
\usepackage{graphicx}
\usepackage{textcomp}
\usepackage{xcolor}
\def\BibTeX{{\rm B\kern-.05em{\sc i\kern-.025em b}\kern-.08em
    T\kern-.1667em\lower.7ex\hbox{E}\kern-.125emX}}
    
\title{Bilingual Word-Level Language Identification for Omotic Languages}
\titlerunning{Bilingual Word-Level LID for Omotic Languages}

\author{Mesay Gemeda Yigezu \and
Girma Yohannis Bade \and
Atnafu Lambebo Tonja \and
Olga Kolesnikova \and
Grigori Sidorov \and
Alexander Gelbukh}
\authorrunning{M.\,G. Yigezu et al.}

\institute{Instituto Politécnico Nacional (IPN), Centro de Investigación en Computación (CIC), Mexico City, Mexico\\
\email{\{mgemedak2022,sidorov,gelbukh\}@cic.ipn.mx}\\
\email{\{girme2005,atnafuatx,kolesolga\}@gmail.com}
}

\begin{document}
\maketitle
\begin{abstract}
 
Language identification is the task of determining the languages for a given text. In many real-world scenarios, text may contain more than one language, particularly in multilingual communities. Bilingual Language Identification (BLID) is the task of identifying and distinguishing between two languages in a given text. This paper presents BLID for languages spoken in the southern part of Ethiopia, namely Wolaita and Gofa. The presence of words’ similarities and differences between the two languages makes the language identification task challenging. To overcome this challenge, we employed various experiments on various approaches. Then, the combination of the Bert-based pre-trained language model and LSTM approach performed better, with an F1-score of 0.72 on the test set. As a result, the work will be effective in tackling unwanted social media issues and providing a foundation for further research in this area.
\keywords{Keywords: LID \and Wolayta and Gofa language \and Omotic language \and bilingual LID \and language identification}

\end{abstract}

\section{Introduction}
The process of automatically identifying the language(s) present in a document based on the content of the document is known as Language identification(LID) ~\cite{ref_article1,mel27}. It was typically assumed that every text is written for one's known languages  ~\cite {ref_article2,mel28}. The task of identifying the most similar languages from the set of bilingual languages data sets is possible through LID approaches. 

Over the past few decades, a fair amount of research has been done on the challenge of determining the language in which a certain text was written. In both academic and professional settings, automated methods for identifying written languages are frequently applied to spoken and written source materials. Despite this universal acceptance, from the point of ~\cite{ref_article3,mel29}, the research in the identification of written languages indicates a number of unanswered topics that are ready for additional study. 
One of the crucial steps in a natural language processing (NLP) challenge is LID. Because it might attempt to predict the  natural language's texts to identify one language from another. Prior to working on any NLP tasks such as sentiment analysis or translation, it is important to understand the language of the text. For instance, the Google Translator application has a menu called "Detect Language", because before Google translates a sentence, it must first determine the language~\cite{ref_article4,mel30}.

Now a day many social media users leave their ideas with a small text, such as a tweet, headline, or comment on various issues but may not have enough information to identify the language accurately and may not be recognized or supported by LID tools.  A single sentence can belong to more than one language, and this can cause difficulties in LID ~\cite{ref_article5,mel31}. 

From the best knowledge of researchers, there is no work conducted on LID for the Omotic language family of Ethiopian languages. Since specially, Ethiopian Omotic language groups  use Latin scripts to write their languages and it's quite similar for those all ~\cite{ref_article6,mel32}, identifying one's language from others needs giving high attention to catch up a little difference that might come from spoken accents.
Therefore in this paper, we focus on bilingual LID for highly interrelated and geographically bordered Omotic languages(Wolayta and Gofa) which are spoken in the southern part of Ethiopia. This study has the following contributions and improvements:
\begin{itemize}
    \item We have prepared a word-level open source dataset for two languages for this study.
    \item The proposed model will be applied in the other low-resource Ethiopian languages.
    \item Social medial issues like hate speech, spreading of rumours, conspiracy theories,  racism, and prejudice speakers will be identified where is written from.
    \item Improved communication:- LID can make it easier for people to communicate with others who speak different languages, reducing language barriers. This means that when someone speaks or writes in a particular language, the system can quickly identify it without requiring the user to specify the language manually. 
    \item Enhanced customer Service:- by identifying multiple languages, businesses can offer better customer service and support to a wider range of customers.
    \item Increased accessibility:- with the ability to recognize multiple languages, technology, and applications can be made more accessible to a wider audience.
\end{itemize}


\section{Related Work}

In recent years, there has been a shift towards using deep learning(DL) approaches for bilingual word-level LID. Sun et al. ~\cite{ref_article7,mel33} proposed using a convolutional neural network (CNN) to identify the language of a given word. They found that CNN performed well in identifying the language of words in the bilingual text and outperformed traditional approaches such as Hidden Markov Models.

Yigezu et al. ~\cite{ref_article8,mel34} proposed Bidirectional Long Short Term Memory (BiLSTM) with attention to identifying code-mixed English-Kannada language. The authors used a CoLI-Kanglish data set which contains 14,847 and 4,585 words for training and testing respectively. In addition to BiLSTM, the authors also employed different ML techniques for the same data sets. The performance evaluation result of the model developed with the deep learning approach gained a macro F1-score of 0.61 and which is better than that of a model developed with ML techniques. As a strength, the work outlined the annotation of data sets well, and it employed different text classification approaches like ML, DL, and combined ones. When it was compared with ~\cite{ref_article9,mel35} text classification work, deep learning-based  systems less perform than transformer-based approaches. 

Barman et el. ~\cite{ref_article10,mel36} used three approaches to identify trilingual languages such as Bengali, English, and Hindi. The proposed approaches are unsupervised dictionary-based approach, supervised word-level classification, and sequential labeling using random fields. Finally, the Unsupervised dictionary-based has performed better results on the given data sets. Similarly, Amine et al.~\cite{ref_article11} state that Unsupervised multilingual text recognition can analyze vast amounts of data fast and does not require any prior knowledge (classes, initial partition). It has the ability to only visualize the findings. 

Tonja et al. ~\cite{ref_article12,mel37} proposed transformer-based word LID for code-mixed English-Kannada language. The authors combined a transformer-based pre-trained language model with a deep learning approach. The authors used BERT \footnote{https://huggingface.co/bert-base-uncased} as the embedding layer and the LSTM algorithm for the classification of the language to assigned labels for LID. Their proposed model better outperformed other pre-trained language models.

Overall, there have been many different approaches proposed for bilingual word-level LID with varying levels of success. However, recent deep learning approaches have shown promising results and may be a promising direction for future  text classification ~\cite{ref_article13}.

\section{Languages}
\subsection{Ethiopian Language Overview}
\begin{figure}[h!]
    \centering
\includegraphics[width=18cm,height=6cm,keepaspectratio]{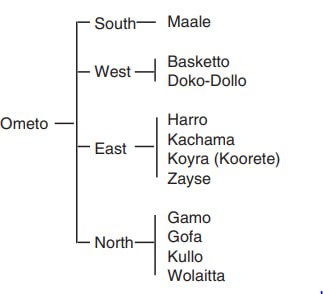}
    \caption{Overview for classification of Omotic languages}
    \label{fig:alpha1}
\end{figure}

Ethiopia is home to a diverse range of languages, with over 83 distinct languages and up to 200 spoken dialects ~\cite{ref_article14}~\cite{ref_article15}. These languages are grouped into four main families: Semitic, Cushitic, Omotic, and Nilo-Saharan. The Omotic language family is one of six within the Afro-Asiatic phylum Figure ~\ref{fig:alpha1} and is primarily spoken in the southern rift valley and around the Omo River ~\cite{ref_article16},~\cite{ref_article17}. Within the Omotic branch, the Ometo languages form a large group that shares similar phonology, grammar, and vocabulary, making them distinct genetic units. Ometic languages are divided into four categories: North, South, East, and West Omotic. Wolayta and Gofa languages are grouped under Northern Omotic ~\cite{ref_article18}.

\subsection{Wolayta Language}

\begin{figure}[h!]
    \centering
    \includegraphics[width=20cm,height=7cm,keepaspectratio]{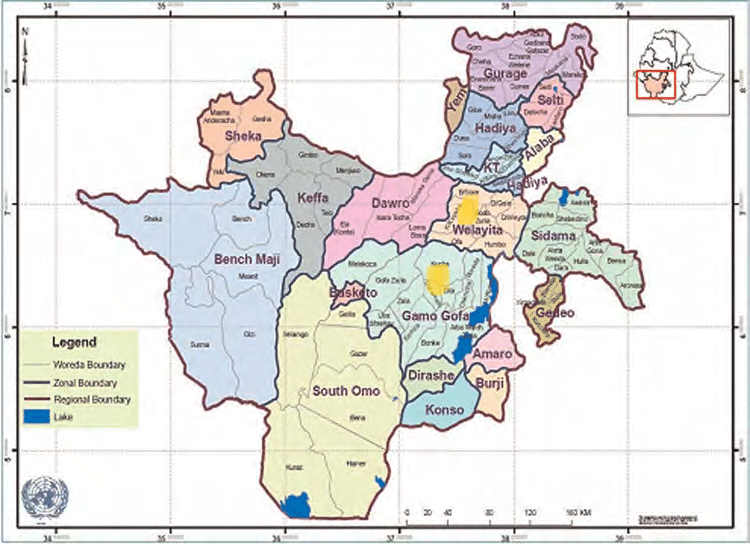}
    \caption{ Southern region of Ethiopia map ~\cite{ref_article20}} 
    \label{fig:alpha}
\end{figure}

Wolayta is a north Omotic language spoken by the Wolayta people in southern Ethiopia ~\cite{ref_article15}, ~\cite{ref_article19}. It is a member of the Omotic branch of the Afro-Asiatic language family. Wolayta is spoken by approximately 2.48 million people, primarily in the Wolayta Zone and some parts of Ethiopia's Southern Nations, Nationalities, and Peoples' Region (SNNPR). It is also spoken by a smaller number of people in neighbouring Zones and Regions such as Dawuro, Gofa, Sidama, and Gamo. Wolayta is one of the most widely spoken Omotic languages and is an official language in the Wolayta Zone.
Figure \ref{fig:alpha} depicts the geographical location of the Wolayta and Gofa zone(yellow-shaded) ~\cite{ref_article20}.

\subsection{Gofa Language}

The term "Gofa" encompasses the language, culture, and region of southern Ethiopia, spoken by the inhabitants of the Gofa Zone and neighbouring communities. With over 362,000 speakers, it is used as the primary language of instruction in primary schools and is also taught as a subject in secondary and high schools. Additionally, it serves as the main language of communication in government offices within the Gofa Zone ~\cite{ref_article21}.

\subsection{Challenges of languages}
The study ~\cite{ref_article10} says "automatic LID on social media data becomes a necessary and challenging task." Particularly, the challenges of Omotic languages are the irregularity of the words, stop word identification issues, compounding, and languages' limited access to digital data, according to ~\cite{ref_article6}. Wolayta and Gofa languages are members of the Omotic languages that are spoken around river Omo, the southern part of Ethiopia.
LID is still a difficult subject to solve, especially when trying to distinguish between languages that are quite similar ~\cite{ref_article22}. 
A number of factors influenced the performance of BLID, including the similarity of the languages being identified and the length of the text. For example, the proposed languages that are more similar to each other are typically more difficult to distinguish than those that are less similar. Similarly, the longer text is typically more reliable for BLID than the shorter one, as it provides more information about the language being used. Further research is needed to continue to improve the performance of BLID methods and to explore new applications for these methods.
\begin{table}[]
\centering
\caption{sample words/phrases in English, Wolayta, Gofa, Wolayta-Gofa and its remark as similar/common or different in written text.}
\begin{tabular}{|c|c|c|c|c|c|}
\hline
No                 & In English            & In Wolayta       & In Gofa        & \begin{tabular}[c]{@{}c@{}}Wolayta-Gofa \\ (common)\end{tabular} & Remark                               \\ \hline
\multirow{4}{*}{1} & While following       & Kaallidi         & Kaallidi       & Kaallidi                                                         & \multirow{4}{*}{Use common words for both}     \\ \cline{2-5}
                   & Country               & Biittaa          & Biittaa        & Biittaa                                                          &                                      \\ \cline{2-5}
                   & Bad                   & Iita             & Iita           & Iita                                                             &                                      \\ \cline{2-5}
                   & Many                  & Daro             & Daro           & Daro                                                             &                                      \\ \hline
\multirow{4}{*}{2} & From his/her/its talk & Haasayaappe      & Hayssafe       & -                                                                & \multirow{4}{*}{Use different but closest words} \\ \cline{2-5}
                   & While I writing      & Xaafayida dayiis & Xaafettidayssi & -                                                                &                                      \\ \cline{2-5}
                   & Are you going?        & Bayii?           & Biyey?         & -                                                                &                                      \\ \cline{2-5}
                   & His/her born          & Yeletay          & Yeletethay     & -                                                                &                                      \\ \hline
\end{tabular}
\label{tab22}
\end{table}

Table \ref{tab22} shows us the similarity and differences between both languages. The first, No column shows the similarity between Wolayta and Gofa words, and the second No column, shows their difference. The differences came from the accents of the users and the similarities came from due to the two language speakers' geographical location in the same area and their languages belonging to the same group, thus the structure and morphology of both languages are quite similar. Thus these circumstances made it difficult to identify the two languages one form other.

\section{Methodology}
We followed different methods such as dataset collection, pre-processing, annotation, approach selection, parameter tuning, conducting experiments and so on. 
\subsection{Dataset development}

For this study, we collected a novel dataset from religious, educational, and social media sources. For the religious domain, we used the data from ~\cite{ref_article23}, for the educational domain, we collected dataset from Wolayta Sodo University, Wolayta language department\footnote{ftp://ftp.wsu.edu.et/ } and for social media domain, we collected from Facebook user comments. 

\subsection{Data pre-processing}

To maintain the model quality, the data pre-processing task must be carried out before the experiment. Pre-processing is the process of cleaning and making data suitable for execution ~\cite{ref_article24} -~\cite{ref_article24b}. In order to generate a high-quality training dataset, we pre-processed the data that was acquired using a variety of different methods. During the process, we carried out a variety of tasks, such as removing any URLs, non-alphabetic characters, numbers, marks, and HTML tags and converting text into lowercase. Then we compiled the pre-processed dataset into a single file and shuffled it. We used Excel Kutools \footnote{https://es.extendoffice.com/product/kutools-for-excel.html} to remove the matching common words from both Wolayta and Gofa to annotate the data.

\subsection{Data annotation}

Before starting the annotation of the data set, deciding what specific annotating tasks should be performed is necessary. This could be labeling data and tagging entities.  
The annotation indicates whether the selected word is spoken in either Wolayta, Gofa, or both for each language. The compiled dataset is manually normalized and annotated with LID. Annotation tasks are performed as mentioned in the LID. The shuffled data is given to annotators of each language. These examples are presented in Table \ref{tab1} which shows how the annotators annotated the training datasets.

LID refers to the process of automatically determining the language in which a given piece of text is written. The task of LID is an important component in (NLP). Its primary goal is to classify the input data into specific language categories from a predefined set of languages. The LID classify into three categories: wal (Wolayta), gof (Gofa) and wal-gof (Wolayta-Gofa).

\textbf{wal:} pertains to the Wolayta language used by the Wolayta people.

\textbf{gof:} denotes the Gofa language spoken by the Gamo and Gofa people. 

\textbf{wal-gof:} is a LID term utilized when both the Wolayta and Gofa people communicate, and it indicates the presence of shared words or linguistic elements during their interactions.
\begin{table}[]
\caption{Sample words with their tags and their English meanings.}
\centering
\begin{tabular}{|l|l|l|}
\hline
\textbf{Word}             & \textbf{Tag}     & \textbf{Translated in English} \\ \hline
Hayassafe        & gof     & don’t speak           \\ \hline
Xaafettidayssi   & gof     & while I’m writing     \\ \hline
Hintte           & gof     & You all               \\ \hline
Asa              & wal     & People                \\ \hline
Giddiis          & wal     & Enough                \\ \hline
kaalletanawu     & wal     & To lead               \\ \hline
Kaallidi         & wal-gof & While following       \\ \hline
Hara             & wal-gof & Other                 \\ \hline
Eridi            & wal-gof & knowingly             \\ \hline
Doonan           & wal-gof & His/her language      \\ \hline
\end{tabular}
\label{tab1}
\end{table}
\subsection{Annotation procedure}
In this task, we recruited three native speaker annotators for each language and the criteria to be selected as an annotator is a person who speaks the chosen language and knows about its structure and morphology.
We held multiple one-on-one sessions with the annotators to thoroughly go over the annotation guidelines. The dataset was split into two batches, and once each batch was annotated, the authors of the paper reviewed the labeled samples. Throughout the annotation process, annotators were instructed to carefully read each text and initially determine the appropriate label for the LID in the text. Subsequently, they were asked to assign the most fitting LID based on the predefined categories.
Finally, we collected the annotated data and calculated the number of language IDs in each word, and we decided on a single label based on a majority vote. when three annotators are involved in the language identification task, the minimum requirement for a language ID to be selected is that at least two out of the three annotators must agree and assign the same language label to a given text. If there is a consensus among two annotators for a specific language ID, then that ID is considered valid and eligible for further consideration or analysis. This ensures a level of agreement and consistency among the annotators in the language identification process.

\subsection{Dataset statistics}
\begin{figure}[!h]
\centering
\includegraphics[width=9cm,height=7cm,keepaspectratio]{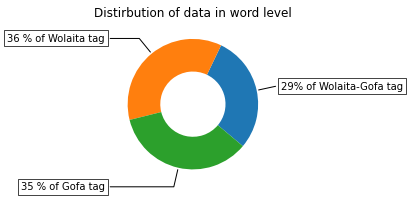}
    \caption{Distribution of tags among whole data sets(in \%)}
    \label{fig:alpha5}
\end{figure}

After being data annotated we collected a total of 144K words from the above three domains and tagged them as Wolayta, Gofa, and wal-gof. Figure \ref{fig:alpha5}  shows the distribution of the collected data with their respective language tags.

\section{Design of proposed Model}
\subsection{Proposed Experimental Architecture}
The proposed architecture is designed to demonstrate how the model identifies bilingual languages. The main phases of the architecture include:\\
\textbf{Phase-1} - creating labeled data for training, the data set contains words together with their respective tags.\\
\textbf{Phase-2} - preprocessing such as tokenization, removing irrelevant data.\\
\textbf{Phase-3} - the tags are converted into machine-readable form .\\
\textbf{Phase-4} - after the label encoding process is complete, the representation for each token is transferred to ML and DL then evaluates the trained model. Also, each token is transferred to transformer layers which allow a pre-trained language model to produce contextualized tokens.\\
\textbf{Phase-5} - the embeddings that were obtained in step 4, were then fed into an LSTM model in order to generate the language tag that corresponds to them and finally evaluate the trained model.\\
\textbf{Phase-6} - the trained model was evaluated by accuracy metrics.\\
\textbf{Phase-7} - after being evaluated, the output which is the predictive model is developed.\\

\begin{figure}[h!]
    \centering
    \includegraphics[width=14cm,height=17cm,keepaspectratio]{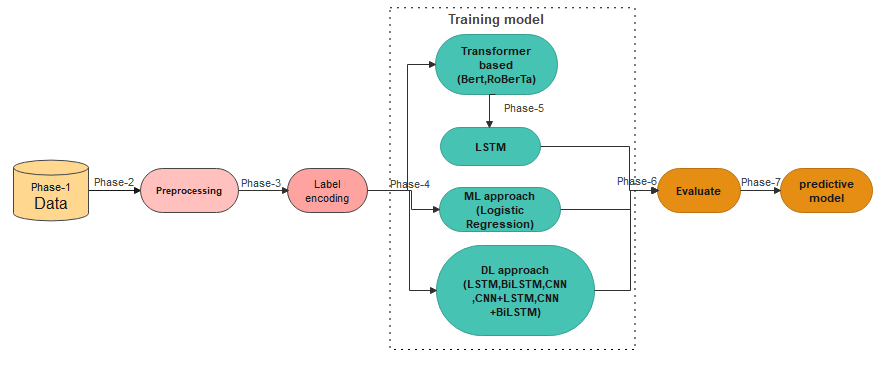}
    \caption{Architecture of proposed model } 
    \label{fig:arch}
\end{figure}

\subsection{Experiment and parameter setting}
\subsubsection{Experiment}

In this study, we present an experimental evaluation of several approaches to BLID. We developed the proposed model using the following approaches:\\
\textbf{Machine learning} is a branch of artificial intelligence that focuses on developing algorithms and statistical models that enable computers to automatically improve their performance on a given task through experience ~\cite{ref_article25}.\\
\textbf{Deep learning} is a subfield of machine learning that uses algorithms inspired by the structure and function of the brain, known as artificial neural networks, to process and analyze large amounts of data ~\cite{ref_article26} -~\cite{ref_article26b}.\\
\textbf{Transformer based} is a deep learning model based on neural networks, specifically designed for processing sequences of data, such as natural language text ~\cite{ref_article27} -~\cite{ref_article27a}.  

In order to train the algorithms and develop the model, we have used various platforms. Among those, the basics are mentioned below.  We have used a cloud-based platform known as Google Colab. \footnote{https://colab.research.google.com/} which is
a popular open-source framework for deep learning with rich libraries such as TensorFlow \footnote {https://tensorflow.org/}, PyTorch \footnote {https://pytorch.org/}, and Keras \footnote {https://keras.io/}.

\subsubsection{parameter setting}
It's important that the optimal parameter settings for a BLID model depend on the specific iterative problem and may require some trial and error to find the best settings. 
When we train a model using a deep learning approach for BLID, there are several parameters that must be set in order to optimize the performance of the model. \\ 
For the purpose of defining the number of times that the learning algorithm will run through the full training data set, with a range of 10 to 30 epochs. 
We used the Leaky ReLU function, which allows returning a small positive value to "leak" through for negative input values while using standard relu 0 for all negative input values. In addition to that, the softmax function is used in the output layer of a classifier, such as a neural network, to convert the raw output values into probabilities. This is useful in multi-class classification problems where there are multiple possible outcomes. \\
Furthermore, we applied batch normalization to the input of each layer, which helped stabilize the training process and improve the performance of the network. It is typically applied to the input of each layer before the activation function, and it is typically used in convolutional neural networks and fully connected networks. We adjusted the batch size to 128 in the model configuration to impact the speed and accuracy of the training process.\\ 


We utilized the learning rate amount of 0.001, which helps to control the step size at which the algorithm updates the weights and biases of the neural network during the training process. 
This is done by iteratively adjusting the parameters of the model and evaluating the error until the minimum error is found. We applied the Adam optimizer in deep learning to find the optimal parameters for a model.\\
Generally, in this experiment, we experimented with four categories. 

The experiment was trained on given datasets with three tags, allowing them to learn the distinctive features of each language and identify them accurately. In the first category, we applied the machine-learning technique of logistic regression. \\
In the second category, to evaluate the performance of the LID model, we used a deep learning approach such as LSTM, BiLSTM, and CNN. After pre-processing our textual data, The text was tokenized and trained with the mentioned approach with the help of a trained data set.\\
In the third experiment category, we evaluated our model by using a combination of CNN with LSTM and BiLSTM. To extract features for the given text, we used CNN and then used LSTM and BiLSTM as classifiers.\\
In the fourth experiment, We used transformer-based models such as BERT-based uncased and RoBERTa-based to develop our model. Finally, We evaluated the performance of our models using several techniques such as accuracy, precision, recall, and F1-score. We discussed our results in Section 6.

\section {Results and Discussion} 

The proposed models are trained on given words in both languages with three tags, allowing them to learn the distinctive features of each language and identify them accurately. Table \ref{tab33} shows the experimental results

\begin{table}[]
\centering
\begin{tabular}{|c|c|c|c|c|c|}
\hline
Experiment             & Approaches                         & Model                    & Precision & Recall & F1-score \\ \hline
Exp-1                  & Machine Learning(ML)               & Logistic Regression      & 0.55      & 0.76   & 0.47     \\ \hline
\multirow{3}{*}{Exp-2} & \multirow{3}{*}{Deep learning(DL)} & LSTM with attention      & 0.78      & 0.43   & 0.55     \\ \cline{3-6} 
                       &                                    & BiLSTM wiith attention   & 0.7       & 0.66   & 0.62     \\ \cline{3-6} 
                       &                                    & CNN                      & 0.77      & 0.55   & 0.64     \\ \hline
\multirow{2}{*}{Exp-3} & \multirow{2}{*}{Combination of DL} & CNN + LSTM               & 0.69      & 0.64   & 0.66     \\ \cline{3-6} 
                       &                                    & CNN + BiLSTM             & 0.71      & 0.67   & 0.69     \\ \hline
\multirow{2}{*}{Exp-4} & \multirow{2}{*}{Combination of PLM and DL} &   Roberta base + LSTM      & 0.77      & 0.64   & 0.69     \\ \cline{3-6} 
                       &                                &   Bert-base-uncased + LSTM & \textbf{0.76}      & \textbf{0.69}   & \textbf{0.72}    \\ \hline
\end{tabular}
\caption{Results for the experiments}
\label{tab33}
\end{table}

In the first experiment, we perform an ML technique which is Logistic Regression and achieves 0.47 with F1-score.\\
In the second experiment, we explored three DL techniques for performing LID at the word level in the bilingual language. We used character-level LSTM and BiLSTM with attention and CNN techniques to perform and showed different performance outcomes based on the same dataset being used. The results showed that the BiLSTM with attention model performed well in identifying the language of text, with an F1-score of 0.68. \\
Thirdly, the experiment was conducted with a combination of LSTM, BiLSTM, and CNN. In this experiment, we used CNN for feature extraction. By using convolutional and max-pooling layers, the model was able to capture the features of the text. In that experiment, the combination of BiLSTM and CNN performed better, with an F1-score of 0.69.\\
The fourth experiment indicates that the transformer-based model has been found to perform better than an entire experiment. We applied a pre-trained language model, which is Bert uncased in the base layer and Roberta in the embedding layer, with the LSTM model. The Bert base uncased model outperforms the Roberta base pre-trained language models that were used in the embedding layer with the LSTM model, with a macro score of 0.76 precision, 0.69 recall, and 0.72 F1-scores. 

The model is based on the LSTM architecture, which is a type of RNN that can capture long-term dependencies in sequential data. The model computes the embeddings of the input text using a pre-trained model. The embeddings are a dense representation of the input text that can capture semantic and syntactic information. The embeddings are fed to an LSTM layer with 128 units. The output of the LSTM layer is passed through a batch normalization layer, which normalizes the activations of the previous layer. The normalized activations are then fed to a dense layer with 768 units, followed by a ReLU activation function. Another dense layer with 768 units and a dropout layer with a dropout rate of 0.1 is applied next. Finally, the output of the dropout layer is passed through a dense layer with 3 units and a softmax activation function, which produces the output probabilities for each of the 3 classes.

Therefore, this model is able to effectively learn the language patterns in the bilingual text dataset. This allowed the model to effectively distinguish between different languages even when the text was short and had limited information.
\section{Conclusion}

The experiments conducted on bilingual LID at the word level using various machine learning, deep learning techniques, and transformer-based model, specifically the Bert base uncased model in the embedding layer of LSTM, performed the best with a 0.72 F1-score among the other models tested. The combination of BiLSTM and CNN also showed good performance but was slightly outperformed by the Bert base uncased model. These results suggest that using pre-trained language models such as Bert and Roberta can be an effective approach for identifying the language of bilingual text.



This study can be used as a starting point for further research and development in this area. Additionally, this study also highlights the importance of creating and annotating datasets for BLID. 

To improve the performance of the bilingual LID model, further research can focus on developing more advanced algorithms, increasing the size ~\cite{ref_article28} and diversity of the training data, and experimenting with different feature representations. Additionally, an important application of the bilingual LID model lies in machine translation tasks. By accurately detecting the source language, the appropriate language model for translation can be selected. This way, the translation system can apply language-specific rules, context, and vocabulary, resulting in more precise and contextually relevant translations ~\cite{ref_article29}.\\

\section{Acknowledgments}
The work was done with partial support from the Mexican Government through the grant A1-S-47854 of CONACYT, Mexico,
grants 20241816, 20241819,  and 20240951 of the Secretaría de Investigación y Posgrado of the Instituto Politécnico Nacional, Mexico. The authors thank the CONACYT for the computing resources brought to them through the Plataforma de Aprendizaje Profundo para Tecnologías del Lenguaje of the Laboratorio de Supercómputo of the INAOE, Mexico and acknowledge the support of Microsoft through the Microsoft Latin America PhD Award.

\end{document}